%% file: acl_latex.tex
\documentclass[11pt]{article}

\PassOptionsToPackage{authoryear,sort}{natbib}
\usepackage[final]{acl}

\usepackage{times}
\usepackage{latexsym}

\usepackage[T1]{fontenc}

\usepackage[utf8]{inputenc}

\usepackage{microtype}

\usepackage{inconsolata}

\usepackage{graphicx}

\usepackage{float}

\usepackage{amsmath}
\usepackage{amssymb}
\usepackage{amsthm}
\usepackage{algorithm}
\usepackage{algorithmic}
\usepackage{booktabs}
\usepackage{multirow}
\usepackage[table]{xcolor}  
\usepackage{url}
\usepackage{enumitem}  
\usepackage{mdframed}  
\usepackage[most]{tcolorbox}  
\usepackage{mathtools}

\usepackage{geometry}

\newcommand\blfootnote[1]{%
  \begingroup
  \renewcommand\thefootnote{}\footnote{#1}%
  \addtocounter{footnote}{-1}%
  \endgroup
}

\title{TiMem: Temporal-Hierarchical Memory Consolidation\\ for Long-Horizon Conversational Agents}

\author{
  \textbf{Kai Li\textsuperscript{1,2,3\dag}},
  \textbf{Xuanqing Yu\textsuperscript{1,2,3\dag}},
  \textbf{Ziyi Ni\textsuperscript{1,2}},
  \textbf{Yi Zeng\textsuperscript{4}},
  \textbf{Yao Xu\textsuperscript{1,5}},
  \textbf{Zheqing Zhang\textsuperscript{6}} \\
  \textbf{Xin Li\textsuperscript{7,8}},
  \textbf{Jitao Sang\textsuperscript{9}},
  \textbf{Xiaogang Duan\textsuperscript{10}},
  \textbf{Xuelei Wang\textsuperscript{1,2*}},
  \textbf{Chengbao Liu\textsuperscript{1,2*}},
  \textbf{Jie Tan\textsuperscript{1,2}} \\
  \footnotesize
  \textsuperscript{1}\textit{Institute of Automation, CAS}\quad
  \textsuperscript{2}\textit{School of Artificial Intelligence, UCAS}\quad
  \textsuperscript{3}\textit{AI Lab, AIGility Cloud Innovation} \\
  \footnotesize
  \textsuperscript{4}\textit{North China Electric Power University}\quad
  \textsuperscript{5}\textit{Beijing Academy of Artificial Intelligence}\quad
  \textsuperscript{6}\textit{Gaoling School of Artificial Intelligence, RUC} \\
  \footnotesize
  \textsuperscript{7}\textit{School of Biomedical Engineering, USTC}\quad
  \textsuperscript{8}\textit{Suzhou Institute for Advance Research, USTC} \\
  \footnotesize
  \textsuperscript{9}\textit{School of Computer Science and Technology, BJTU}\quad
  \textsuperscript{10}\textit{Hunan Central South Intelligent Equipment Co., Ltd.} \\
  \footnotesize
  \texttt{\{likai2024, yuxuanqing2021, niziyi2021, xuelei.wang, liuchengbao2016, jie.tan\}@ia.ac.cn, 120231290122@ncepu.edu.cn} \\
  \footnotesize
  \texttt{yao.xu@nlpr.ia.ac.cn, zhangzheqing@ruc.edu.cn, merley@mail.ustc.edu.cn, jtsang@bjtu.edu.cn, Duanxg@zeqp.net}\\
}

\begin{document}

\maketitle

\blfootnote{\textsuperscript{\dag}These authors contributed equally to this work.}
\blfootnote{\textsuperscript{*}Corresponding authors.}

\begin{abstract}

Long-horizon conversational agents have to manage ever-growing interaction histories that quickly exceed the finite context windows of large language models (LLMs). Existing memory frameworks provide limited support for temporally structured information across hierarchical levels, often leading to fragmented memories and unstable long-horizon personalization.
We present \textbf{TiMem}, a temporal--hierarchical memory framework that organizes conversations through a Temporal Memory Tree (TMT), enabling systematic memory consolidation from raw conversational observations to progressively abstracted persona representations.
TiMem is characterized by three core properties: (1) temporal--hierarchical organization through TMT; (2) semantic-guided consolidation that enables memory integration across hierarchical levels without fine-tuning; and (3) complexity-aware memory recall that balances precision and efficiency across queries of varying complexity.
Under a consistent evaluation setup, TiMem achieves state-of-the-art accuracy on both benchmarks, reaching 75.30\% on LoCoMo and 76.88\% on LongMemEval-S. It outperforms all evaluated baselines while reducing the recalled memory length by 52.20\% on LoCoMo. Manifold analysis indicates clear persona separation on LoCoMo and reduced dispersion on LongMemEval-S.
Overall, TiMem treats temporal continuity as a first-class organizing principle for long-horizon memory in conversational agents. The code is available at \url{https://github.com/TiMEM-AI/timem}.

\end{abstract}

\section{Introduction}
\label{sec:introduction}
\vspace{-0.4em}

\begin{figure}[!t]
\vspace{1.5em}
\centering

\includegraphics[width=\columnwidth]{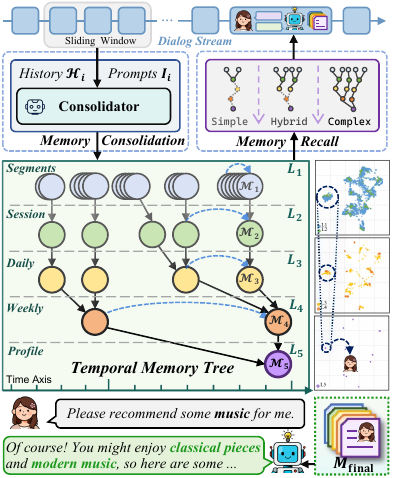}
\caption{TiMem framework overview. The framework organizes conversational streams through a five-level TMT, consolidating memories from factual segments to persona profiles, with adaptive memory recall guided by query complexity.}
\label{fig:timem-overview}
\end{figure}

Large Language Models (LLMs) have enabled conversational agents to evolve from short-horizon task solvers \cite{qian-etal-2024-chatdev,zhang-etal-2024-android,zeng-etal-2024-agenttuning} to long-horizon personalized companions \cite{zhang2024personalization,li-etal-2025-hello,chen2024personapersonalizationsurveyroleplaying}. Supporting such interactions requires two capabilities: maintaining temporal coherence as user states evolve, and forming stable representations by distilling consistent personas from dynamic experiences. However, interaction histories grow unbounded, while LLMs operate under finite context windows, making it harder to sustain temporally consistent personalization at scale. The key challenge is to transform long-horizon experience into compact representations that remain temporally grounded and useful for subsequent tasks.

Existing solutions under-emphasize temporal structure as a first-class constraint, and often lack explicit temporal containment guarantees across hierarchical levels. Parametric approaches expand context windows~\cite{gemini15,rope-scaling} or optimize internal context capacity~\cite{bini2025memloradistillingexpertadapters,bui2025cachelaststokenretention}, but they remain bounded by model architecture and do not provide persistent cross-session storage. External memory systems~\cite{memgpt,mem0,memorybank} enable persistence but often rely on semantic similarity-driven clustering~\cite{raptor} or learned routing policies~\cite{du2025memr3memoryretrievalreflective}, treating temporal structure as auxiliary metadata. As a result, memories from different periods can be aggregated without clear temporal boundaries, and retrieval may surface temporally distant evidence without an explicit ordering. For evolving users, persona modeling benefits from a time-ordered evidence chain rather than only semantically similar fragments.

Cognitive neuroscience provides a principled perspective on this problem. Human memory relies on complementary learning systems~\cite{mcclelland1995there}, where \textbf{memory consolidation}~\cite{squire2015memory} progressively transforms rapid episodic encoding into more stable semantic structures~\cite{cowan2021memory}. This adaptive process prioritizes goal-relevant information over indiscriminate retention. Translating this view to long-horizon agents suggests two design requirements: time should be encoded as an explicit structural constraint, and memory should be consolidated progressively across temporal granularities.

To this end, we introduce \textbf{TiMem}, a memory framework that uses temporal structure as the primary organizing principle and operationalizes consolidation in a computational form. TiMem consolidates fine-grained episodic interactions into higher-level semantic patterns and persona representations, rather than maintaining raw context buffers.

As illustrated in Figure~\ref{fig:timem-overview}, TiMem implements a hierarchical consolidation mechanism with three components and requires no additional fine-tuning in our experiments. (1) The \textbf{Temporal Memory Tree (TMT)} organizes memories with explicit temporal containment and order through tree constraints. (2) The \textbf{Memory Consolidator} performs instruction-guided consolidation; level-specific prompts control the abstraction level, enabling plug-and-play use across different LLM backends. (3) \textbf{Memory Recall} performs complexity-aware hierarchical retrieval: a recall planner selects appropriate hierarchy levels based on query complexity, and a recall gating step filters candidates to balance factual detail with higher-level personalization.

Our contributions are threefold:
(1) the TMT, a novel structure that enforces explicit temporal containment and granularity for memory organization;
(2) the TiMem framework, a temporal--hierarchical memory consolidation framework based on instruction-guided reasoning and complexity-adaptive recall, requiring no fine-tuning;
(3) a comprehensive evaluation demonstrating TiMem's state-of-the-art accuracy (75.30\% on LoCoMo, 76.88\% on LongMemEval-S) and efficiency (52.20\% reduced recalled context on LoCoMo), with ablations and manifold analyses providing insights into its hierarchical representations.

\begin{figure*}[t!]
\centering
\includegraphics[width=\linewidth]{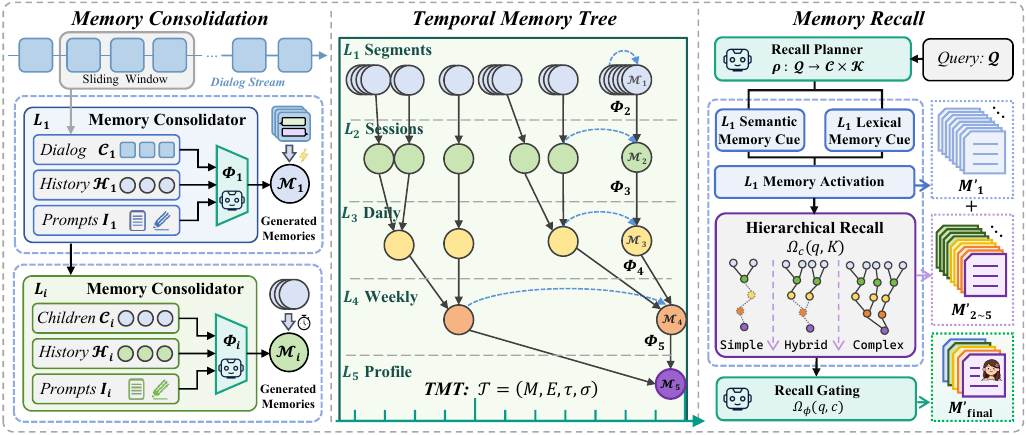}
\caption{TiMem architecture overview: a five-layer TMT from level 1 segments to level 5 profiles, with a consolidation pipeline processing dialog into temporal-hierarchical memories, and a recall pipeline without fine-tuning that includes a recall planner, hierarchical recall, and a recall gating module.}
\label{fig:architecture}
\end{figure*}

\vspace{-0.4em}
\section{Related Work}
\label{sec:related}
\vspace{-0.4em}

\paragraph{Parametric Memory Approaches.}
Context window expansion methods such as Gemini~\cite{gemini15}, LongLoRA~\cite{longlora}, and RoPE scaling~\cite{rope-scaling} alleviate sequence length limits but incur quadratic computational costs and attention dilution~\cite{liu-etal-2024-lost}.
Parametric optimization approaches, including MemLoRA~\cite{bini2025memloradistillingexpertadapters}, HMT~\cite{he2025hmt}, and TRIM-KV~\cite{bui2025cachelaststokenretention}, compress memory through adapter distillation or learned token retention. However, they remain constrained by architectural context windows and do not support persistent cross-session memory.
\vspace{-0.6em}
\paragraph{External Memory Management.}
Semantic clustering approaches, including Mem0~\cite{mem0}, RAPTOR~\cite{raptor}, and MemTree~\cite{rezazadeh2025isolatedconversationshierarchicalschemas}, organize memory through embedding-based similarity aggregation.
Graph-based approaches, including Zep~\cite{zep}, LiCoMemory~\cite{huang2025licomemory}, and Theanine~\cite{ong2025theanine}, explicitly model entity relations and temporal knowledge.
Cognitively motivated frameworks such as A-MEM~\cite{amem}, Nemori~\cite{nemori}, ENGRAM~\cite{patel2025engram}, and RMM~\cite{rmm} employ self-organizing or agentic mechanisms, while preference-aware systems like MemoryBank~\cite{memorybank} and PAMU~\cite{pamu} support personalization through dynamic updates.
OS-inspired memory systems such as MemGPT~\cite{memgpt}, MemoryOS~\cite{memoryos}, and MemOS~\cite{memos} manage long contexts via hierarchical tiers and virtual memory mechanisms.
However, most existing approaches do not treat temporal structure as a first-class organizing principle, resulting in fragmented memory representations and unstable long-horizon behavior.

\vspace{-0.3em}
\section{Methodology}
\label{sec:methodology}
\vspace{-0.4em}

We present \textbf{TiMem}, a temporal--hierarchical memory framework for long-horizon conversational agents. TiMem consists of (i) a TMT that encodes temporal structure, (ii) a Memory Consolidator that performs level-specific consolidation via instruction prompting without fine-tuning, and (iii) a Recall pipeline that uses a planner to select relevant memory levels and a recall gating module to retain query-relevant memories, as illustrated in Figure~\ref{fig:architecture}.

\vspace{-0.6em}
\subsection{Temporal Memory Tree}
\label{subsec:hierarchy}
\vspace{-0.1em}

The TMT provides a stable backbone for long-horizon memory: it preserves temporal coherence, supports progressive consolidation, and reduces noise by transforming details into higher-level abstracts. Lower-level memories cover short intervals and keep concrete details, while higher-level ones span longer intervals and store more consolidated representations. Each node $m$ stores a time interval $\tau(m)$ and a semantic memory $\sigma(m)$. We use $\ell(m) \in \{1,\dots,L\}$ to denote the level of node $m$, from fine-grained to generalized.

\textit{Definition.} TMT is a hierarchical memory structure $\mathcal{T}=(M,E,\tau,\sigma)$ defined by:
\begin{itemize}[leftmargin=*, itemsep=0.2em, topsep=0.2em, parsep=0pt]
  \item $M=\bigcup_{i=1}^{L}M_i$ is the set of memory nodes partitioned across $L$ abstraction levels;
  \item $E\subseteq M\times M$ defines parent--child relationships where $\ell(m_u)=\ell(m_v)+1$, $\forall (m_u,m_v)\in E$;
  \item $\tau$ assigns each node a temporal interval $\tau(m)=[t_{\text{start}}, t_{\text{end}}]$ which is continuous over periods;
  \item $\sigma$ maps each node to a semantic memory $\sigma(m)$ stored as text and embeddings.
\end{itemize}

\textit{Structural Properties.} The structure is governed by three principles that make temporal order explicit and enable progressive abstraction:
\vspace{-0.4em}
\begin{itemize}
    \item \textbf{Temporal Containment}: $\tau(m_u) \supseteq \tau(m_v)$, for each parent-child edge $(m_u,m_v) \in E$, the parent interval covers the child interval.
    \vspace{-0.6em}
    \item \textbf{Progressive Consolidation}: $|M_i| \leq |M_{i-1}|$ ensures higher-level memories are fewer, reflecting consolidation from fine-grained facts to patterns and profiles.
    \vspace{-0.6em}
    \item \textbf{Semantic Consolidation}: Specified by level-specific instruction prompts $\mathcal{I}_i$, $\sigma(m_u) = \text{LLM}(\{\sigma(m_v)\}, \mathcal{I}_i)$ enables hierarchy specialization through the consolidation process.
\end{itemize}
\vspace{-1em}

\paragraph{Implementation} 
TMT supports arbitrary $L$ and $\tau$ configurations. For reproducibility, \textbf{TiMem uses a five-level hierarchy} (segment, session, day, week, profile). Each level performs a different type of consolidation, specified by level-specific instruction prompts $\mathcal{I}_i$:
\vspace{-0.6em}
\begin{itemize}
     \item \textbf{Factual Summarization}: Segments $L_1$ distill key dialog details; Sessions $L_2$ merge into non-redundant event summaries.
     \vspace{-0.6em}
     \item \textbf{Evolving Patterns}: Daily $L_3$ captures routine contexts and recurrent interests; Weekly $L_4$ integrates evolving behavioral features and preference patterns.
     \vspace{-0.6em}
     \item \textbf{Persona Representation}: Profile $L_5$ is an incrementally refined profile capturing stable personality, preferences, and values from long-term patterns, updated on monthly intervals.
\end{itemize}
\vspace{-0.6em}

The framework is designed to be model-independent and does not require fine-tuning; it can be applied across different LLM backbones.

\vspace{-0.6em}
\subsection{Memory Consolidation}
\vspace{-0.2em}
TiMem constructs the hierarchy with a Memory Consolidator that converts dialog into structured memories and uses Stratified Scheduling to balance consolidation efficiency and computational cost.
\vspace{-0.4em}
\subsubsection{Memory Consolidator}
At level $i$, the consolidator generates new memories by prompting an LLM with (i) child memories, (ii) historical memories, and (iii) instruction prompts.
\vspace{-0.2em}
\begin{equation}
     \Phi_i: \mathcal{C}_i \times \mathcal{H}_i \times \mathcal{I}_i \to M_i
     \label{eq:consolidator}
\end{equation}

\vspace{-0.3em}
In formula \eqref{eq:consolidator}, $\mathcal{C}_i$ are child memories from level $i{-}1$, $\mathcal{H}_i$ provides short same-level history for continuity, and $\mathcal{I}_i$ are instruction prompts. We use $\mathcal{I}_1$-$\mathcal{I}_2$ for factual consolidation, $\mathcal{I}_3$-$\mathcal{I}_4$ for pattern consolidation, and $\mathcal{I}_5$ for profile representation. Example consolidator prompt is shown in Appendix~\ref{app:prompt_timem_consolidator}.

\paragraph{Child Memories}
We group the conversation timeline into intervals $g \in \mathcal{G}_i$ (e.g., sessions, days). For $i \geq 2$, child memories for each group are the lower-level nodes whose time spans fall inside $g$:
\vspace{-0.2em}
\begin{equation}
    \mathcal{C}_i(g) = \{ m \in M_{i-1} : \tau(m) \subseteq g \}, \quad i \geq 2
\end{equation}
At the base level ($L_1$), child memories are the raw dialog turns within the interval.
\vspace{-0.2em}
\paragraph{Historical Memories} $\mathcal{H}_i$ consists of the $w_i$ most recent memories from the same level $M_i$:
\vspace{-0.1em}
\begin{equation}
    \mathcal{H}_i = \{ m^{(i)}_{-j} : 1 \leq j \leq w_i \}
\end{equation}
where $m^{(i)}_{-j}$ denotes the $j$-th most recent memory at level $i$. This sliding window provides continuity across temporal groups. We set $w_i=3$ across all levels to ensure consolidation consistency.

\subsubsection{Stratified Scheduling}

Memory consolidation follows a two-tier scheduling strategy that balances freshness and efficiency:
\vspace{-1em}
\begin{itemize}[leftmargin=*, itemsep=0.2em, topsep=0.2em, parsep=0pt]
    \item \textbf{Online consolidation ($L_1$):} Factual segment memories $m^{(1)}_k$ are generated immediately as the dialog progresses. With $w_d=1$ dialog turn (one user--assistant exchange), the consolidator $\Phi_1$ is invoked after each new turn to capture fine-grained evidence.
    \item \textbf{Scheduled consolidation ($L_2$-$L_5$):} Higher-level memories $m^{(i)}(g)$ are generated automatically when their temporal windows end. Upon closure of temporal group $g \in \mathcal{G}_i$, the framework triggers $\Phi_i(\mathcal{C}_i(g), \mathcal{H}_i(g); \mathcal{I}_i)$ to consolidate child memories into a higher-level, more abstract representation.
\end{itemize}

Thus, the stratified design ensures that factual details are captured in real time while consolidation is aligned with predefined temporal boundaries.

\vspace{-0.6em}
\subsection{Memory Recall}

Memory recall traverses the TMT to surface relevant memories, balancing precision, efficiency, and context length. It adapts scope to query complexity: simple questions target exact evidence, while complex ones recall more context across all levels. A final recall gating performs recall-time forgetting, filtering redundancy and conflicts, retaining only memories required for the current interaction.

\subsubsection{Recall Planner}

The planner $p: \mathcal{Q} \to \mathcal{C} \times \mathcal{K}$ maps a query $q$ to a complexity label $c \in \{\text{simple}, \text{hybrid}, \text{complex}\}$ and keywords $K$. We obtain both by prompting an LLM (Appendix~\ref{app:prompt_locomo_qa}), without dataset-specific training or labeled annotations.

Query complexity determines which TMT levels to search. We define three layer groups:
\vspace{-0.6em}
\begin{itemize}
    \item \textbf{Factual Layers} ($\mathcal{L}_{\text{fact}}$): $L_1$-$L_2$ capturing fine-grained event details.
    \vspace{-0.6em}
    \item \textbf{Pattern Layers} ($\mathcal{L}_{\text{patt}}$): $L_3$-$L_4$ behavioral trends and patterns.
    \vspace{-0.6em}
    \item \textbf{Profile Layer} ($\mathcal{L}_{\text{prof}}$): $L_5$ synthesizing long-term, stable characteristics.
\end{itemize}

\vspace{-0.6em}
Although \textit{simple} queries are short, they can still ask about stable preferences, so we include the profile layer $L_5$ by default. Intermediate pattern layers are useful for cross-event reasoning and are therefore emphasized in \textit{hybrid} and \textit{complex} queries.

The recall strategy $\mathcal{S}$ maps complexity $c$ to subsets of TMT:
\vspace{-0.6em}
\begin{align}
\mathcal{S}(\text{simple}) &= \mathcal{L}_{\text{fact}} \cup \mathcal{L}_{\text{prof}} \\
\mathcal{S}(\text{hybrid}) &= \mathcal{L}_{\text{fact}} \cup \mathcal{L}_{\text{patt}}^{\text{partial}} \cup \mathcal{L}_{\text{prof}} \\
\mathcal{S}(\text{complex}) &= \mathcal{L}_{\text{fact}} \cup \mathcal{L}_{\text{patt}} \cup \mathcal{L}_{\text{prof}}
\end{align}
where $\mathcal{L}_{\text{patt}}^{\text{partial}}$ recalls $L_3$ memories, while $\mathcal{L}_{\text{patt}}$ recalls more $L_3$ and $L_4$ memories. Simple queries bypass intermediate consolidated memories by directly accessing factual details and stable profiles, while complex ones traverse the full hierarchy to capture information at all levels.

\subsubsection{Hierarchical Recall}

Hierarchical recall operates in two stages: leaf selection at the base level, followed by hierarchical recall propagation through memory subtrees.

\paragraph{Stage 1: Base-Level Memory Activation}
At $L_1$, dual-channel scoring combines semantic similarity and lexical matching through fusion:
\begin{equation}
\scalebox{0.9}{$s(m,q,K)=\lambda{s}_{\mathrm{sem}}(m,q)+(1-\lambda){s}_{\mathrm{lex}}(m,K)$}
\end{equation}
where ${s}_{\text{sem}}$ is cosine similarity between embeddings, ${s}_{\text{lex}}$ is BM25 score for keyword matching, and $\lambda \in [0,1]$ balances both channels. The top-$k_1$ scoring segments form the leaf set $\Omega_1(q, K)$.

\paragraph{Stage 2: Hierarchical Recall Propagation}
For each leaf $m \in \Omega_1$, we collect its ancestors at the hierarchy levels selected by $\mathcal{S}(c)$:
\vspace{-0.2em}
\begin{equation}
\scalebox{0.95}{$\displaystyle\mathcal{A}(m,c)=\{m'\!\in\! M: m\!\preceq\! m',\ \ell(m')\!\in\!\mathcal{S}(c)\}$}
\end{equation}
where $m \preceq m'$ denotes that $m'$ is an ancestor of $m$, and $\mathcal{S}(c)$ restricts recall to levels specified by complexity $c$. The complete candidate set integrates leaves and their ancestors:
\vspace{-0.4em}
\begin{equation}
\scalebox{0.95}{$\Omega_c(q,K) = \Omega_1(q, K) \cup \bigcup_{m \in \Omega_1(q, K)} \mathcal{A}(m, c)$}
\end{equation}

For brevity, we denote this candidate set as $\Omega_c$. The number of recalled memories per level is determined by query complexity; specific configurations are detailed in Appendix~\ref{app:recall_config}.

\subsubsection{Recall Gating}

Recall gating implements \textbf{recall-time forgetting}: after collecting candidate memories, we keep only the truly useful ones for answering the query.

The recall gating module $\phi$ receives query $q$, its complexity $c$, and the candidate set $\Omega_c$ organized by hierarchy levels. It prompts an LLM to determine whether each memory should be retained:
\begin{equation}
\Omega_\phi(q, c) = \{m \in \Omega_c \mid \phi(m, q, c) = \text{retain}\}
\end{equation}
where $\phi(m, q, c)$ denotes the LLM's retention decision for memory $m$ given query $q$ and complexity $c$. Query complexity guides the breadth of retention: simple queries favor precision by retaining fewer memories, while complex queries favor recall by accepting broader context. Example recall gating prompt template is in Appendix~\ref{app:prompt_timem_relevance_simple}.

 The retained memories are ranked by hierarchy level and temporal proximity within each level:
 \begin{equation}
 \scalebox{0.82}{$\displaystyle\Omega_{\mathrm{final}}(q,c)=\mathrm{sort}\!\big(\Omega_\phi(q,c),\ \mathrm{key}{=}(\ell(m),|t_q{-}t_m|)\big)$}
 \end{equation}
 where $\ell(m)$ denotes the hierarchy level, $t_q$ is the query time, and $t_m=t_{\text{end}}(m)$ so $|t_q - t_m|$ measures temporal distance, organizing relevant memories by recency within each consolidation level, thereby ensuring concise, temporally coherent, and information‑dense responses.
\vspace{-0.4em}
\subsubsection{Recall Pipeline}
\vspace{-0.2em}
The complete recall integrates three stages:

\begin{enumerate}[leftmargin=*, itemsep=0.2em, topsep=0.2em, parsep=0pt]
  \item \textbf{Recall planner}: $p(q)\!\rightarrow\!(c, K)$ predicts complexity $c$ and extracts keywords $K$ to determine the hierarchical search scope.
  \item \textbf{Hierarchical Recall}: Dual-channel scoring selects $L_1$ leaves, then hierarchical recall propagation collects relevant ancestors at planner-specified levels, forming the candidate set $\Omega_c$.
  \item \textbf{Recall Gating}:  The refiner filters the candidates based on query relevance and temporal consistency, then orders them to produce the final memory set $\Omega_{\mathrm{final}}(q, c)$.
\end{enumerate}

This pipeline enables complexity-adaptive recall that balances precision and temporal relevance across TMT's hierarchical structure.

\vspace{-0.6em}
\input{tables/result_r1_locomo.tex}

\input{tables/result_r2_longmemeval.tex}

\vspace{-0.2em}
\section{Experiments}
\label{sec:experiments}
\vspace{-0.4em}

\subsection{Experimental Setup}
\vspace{-0.4em}

\paragraph{Datasets}

We evaluate on two long-term conversational memory benchmarks: \textbf{LoCoMo}~\cite{locomo}, a dataset with 10 user groups across multi-session dialog, and \textbf{LongMemEval-S}~\cite{longmemeval}, including 500 conversations designed for very-long memory processing evaluation.
\vspace{-0.4em}
\paragraph{Baselines}
We compare TiMem with five representative memory baselines using their recommended configurations: \textbf{MemoryBank}~\cite{memorybank}, \textbf{Mem0}~\cite{mem0}, \textbf{A-MEM}~\cite{amem}, \textbf{MemoryOS}~\cite{memoryos}, and \textbf{MemOS}~\cite{memos}.
\vspace{-0.4em}
\paragraph{Implementation Details}

For fair comparison, all methods use the same LLM and embedding setup: \texttt{gpt-4o-mini-2024-07-18} for generation and recall, \texttt{Qwen3-Embedding-0.6B} for embeddings, and recall budget $k\!=\!20$. TiMem uses $\lambda\!=\!0.9$ and $w_i\!=\!3$. We use the LLM-as-a-Judge (LLJ), where an LLM judges answer correctness; we report accuracy along with memory tokens and recall latency for efficiency. Details are in Appendix~\ref{app:implementation}.

\vspace{-0.4em}

\subsection{Main Results}
\vspace{-0.2em}
\subsubsection{Results on LoCoMo}
\vspace{-0.2em}

Table~\ref{tab:locomo_results} shows that TiMem achieves the best overall LLJ accuracy on LoCoMo at 75.30\% ± 0.16\%. It outperforms the strongest evaluated baseline, MemOS, at 69.24\% ± 0.11\%. TiMem also improves F1 and ROUGE-L (RL) to 54.40 and 54.68 in percentage, and achieves the best LLJ score in each question type. We compute LLJ using Mem0's evaluation prompt template, as shown in Appendix~\ref{app:prompt_locomo_judge}.
\vspace{-0.4em}
\subsubsection{Results on LongMemEval-S}

Table~\ref{tab:longmemeval_detailed} shows that TiMem achieves the best overall LLJ accuracy on LongMemEval-S at 76.88\% ± 0.30\% with \texttt{gpt-4o-mini-2024-07-18} as the answer model, outperforming the evaluated baselines. With \texttt{gpt-4o-2024-11-20} as the answer model, TiMem remains best overall at 78.96\% ± 0.26\%. The QA and LLJ protocol follow the official LongMemEval-S evaluation template, as shown in Appendix~\ref{app:prompt_lme_qa} and~\ref{app:prompt_lme_judge}.

\vspace{-0.2em}
\subsection{Ablation Studies}

We ablate TiMem to isolate the contribution of its main components. All ablations use \texttt{gpt-4o-mini-2024-07-18} for LLM operations and \texttt{Qwen3-Embedding-0.6B} for embeddings.
\vspace{-0.4em}
\subsubsection{Planner and Recall Gating}

Table~\ref{tab:ablation_e1} compares seven configurations of recall scope and gating.
Fixed-scope recall under-recalls for Simple queries and introduces noise for Complex ones. Recall gating sharply reduces memory length—for example, from 3710.30 to 367.68 tokens on LoCoMo under Simple—but accuracy drops when the scope is overly narrow. Among fixed-scope settings, Hybrid + Recall Gating performs best, achieving 73.38\% on LoCoMo and 75.00\% on LongMemEval-S. The adaptive planner further improves the accuracy–cost trade-off, reaching 75.30\% with 511.25 tokens on LoCoMo and 76.88\% with 1270.62 tokens on LongMemEval-S.

\vspace{-0.2em}
\subsubsection{Hierarchical Architecture}

Table~\ref{tab:ablation_e2} examines hierarchy depth and recall strategy. With L1-only memories, hierarchical recall propagation raises LongMemEval-S LLJ from 57.40\% to 72.40\% compared to flat recall, indicating that hierarchical propagation recovers necessary temporal dependencies. However, L1-only remains below the full hierarchy on LoCoMo, as isolated factual fragments often lack the broader context required for complex queries. Using only high-level layers (L2--L5) further reduces accuracy, confirming that summaries alone cannot replace fine-grained evidence. Overall, the full hierarchy combines precise L1 grounding with contextual understanding from L2--L5, achieving the best performance on both datasets.

These ablations support TiMem's core design: the temporal hierarchy provides both factual precision and contextual understanding through memory consolidation, while the adaptive planner dynamically balances recall scope.
\vspace{-0.2em}

\input{tables/ablation_e1_routing_filter.tex}

\input{tables/ablation_e2_hierarchy.tex}

\vspace{-0.2em}
\subsection{Memory Manifold Analysis}
\vspace{-0.2em}

\begin{figure}[htbp]
\centering
\includegraphics[width=\columnwidth]{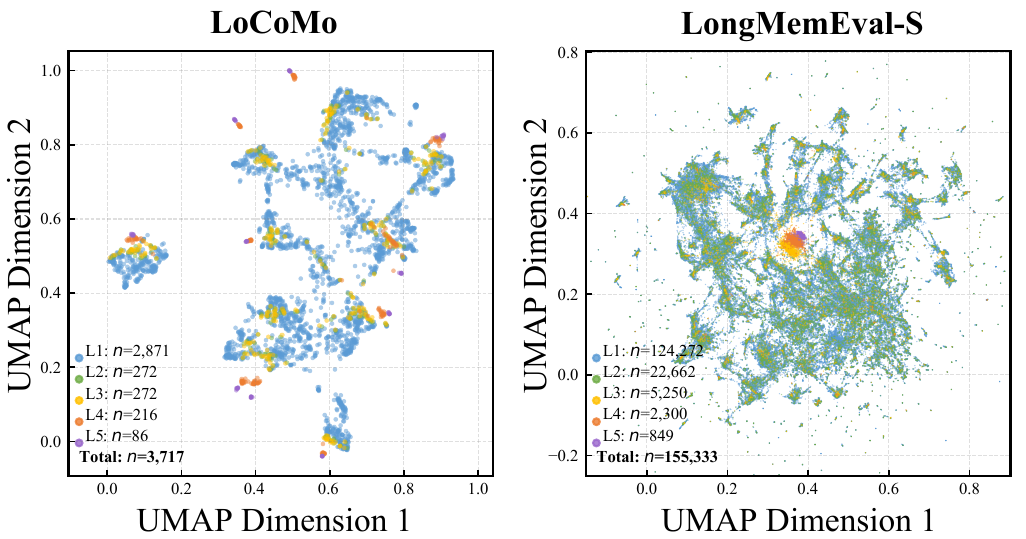}
\caption{UMAP visualization of memory embeddings. \textbf{Left:} LoCoMo exhibits 10 user groups separation through hierarchical consolidation. \textbf{Right:} LongMemEval-S converges toward shared persona structure through noise suppression.}
\label{fig:memory_visualization}
\end{figure}

Figure~\ref{fig:memory_visualization} illustrates UMAP visualization of TiMem memory embeddings on LoCoMo and LongMemEval-S through different hierarchies. It shows that consolidation reshapes memory geometry differently across datasets. On LoCoMo, higher-level memories separate users more clearly, with clustering quality improving 6.2$\times$, indicating effective persona feature distillation. On LongMemEval-S, consolidation reduces spatial dispersion by 50\%, suggesting suppression of sampling noise while retaining core persona attributes. These complementary behaviors suggest that TiMem preserves semantically salient patterns beyond uniform averaging. Detailed metrics are in Appendix~\ref{app:manifold}.

\vspace{-0.2em}
\subsection{Efficiency Analysis}
\vspace{-0.2em}

We evaluate TiMem's efficiency by analyzing both offline consolidation overhead and online recall performance. Table~\ref{tab:consolidation_cost} compares the LLM calls required for consolidation, while Table~\ref{tab:efficiency_metrics} reports memory context length and latency.

\begin{table}[ht]
\centering
\small
\setlength{\tabcolsep}{4pt}
\begin{tabular}{lcccc}
\toprule
\textbf{Dataset} & \textbf{L1-only} & \textbf{Full TMT} & \textbf{Extra Calls} & \textbf{Rate} \\
\midrule
LoCoMo & 2,871 & 3,717 & 846 & +29.5\%  \\
LME-S & 124,272 & 155,333 & 31,061 & +25.0\%  \\
\bottomrule
\end{tabular}
\caption{Amortized consolidation calls comparison between flat L1-only memory and the full five-level TiMem hierarchy. LME-S denotes LongMemEval-S.}
\label{tab:consolidation_cost}
\end{table}

\input{tables/analysis_efficiency_metrics.tex}

As shown in Table~\ref{tab:consolidation_cost}, the full TMT hierarchy increases internal LLM calls by 25\%--30\% compared to a flat L1-only baseline. However, this consolidation cost is amortized over the interaction history and significantly lowers the token cost and latency for the external answering model during inference.

Specifically, on LoCoMo, TiMem recalls only 511.25 tokens per query—a 52.20\% reduction compared to Mem0's 1,070.10 tokens. The P50 recall latency is 2.35s on LoCoMo and 1.76s on LongMemEval-S. While latency is primarily dominated by internal LLM calls for planning and gating, the overall system remains efficient due to the shortened input context. We also observe that context length scales with query complexity, with the more diverse queries in LongMemEval-S requiring broader recall.

\vspace{-0.4em}
\subsection{Parameter Studies}
\vspace{-0.2em}

We conduct a parameter study on LoCoMo.

\textbf{LLM Configuration.} Under the same answering and judgement protocol, TiMem is portable across internal LLMs for memory operations. End-to-end performance is primarily driven by the answering LLM, with the best configuration reaching 80.45\%, indicating that answer-time reasoning dominates once memory quality is adequate.

\textbf{Segment Granularity.} Increasing the L1 segment size consistently degrades accuracy, dropping from 75.30\% at 1 turn to 65.26\% at 8 turns, indicating that finer-grained segments better preserve atomic evidence for downstream QA.

\textbf{Semantic--Lexical Balance.} We investigate the impact of $\lambda$ for hybrid retrieval. As detailed in Appendix~\ref{app:lambda_sensitivity}, TiMem maintains stable performance across $\lambda \in [0.7, 1.0]$, with LoCoMo accuracy ranging from 73.96\% to 75.30\%. Performance peaks at $\lambda = 0.9$, where semantic similarity provides robustness against paraphrasing while lexical matching effectively captures exact entities and rare terms.

Detailed experimental designs, results, and cross-configuration analysis are provided in Appendix~\ref{app:param_studies}.

\vspace{-0.4em}
\subsection{Case Study}
\vspace{-0.3em}

\begin{figure}[htbp]
\centering
\includegraphics[width=\columnwidth]{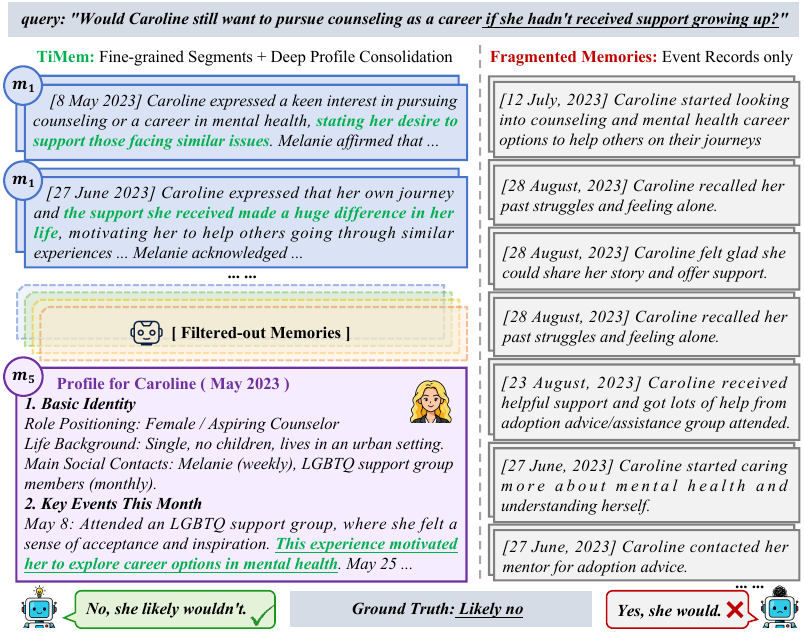}
\caption{Case study comparing TiMem and a non-hierarchical baseline. TiMem’s hierarchical consolidation organizes timestamped evidence into coherent chains and a persona profile, whereas the baseline recalls only isolated event records.}
\label{fig:case_study}
\end{figure}

Figure~\ref{fig:case_study} contrasts TiMem's hierarchical consolidation against Mem0 fragmented memories.

\textbf{TiMem} recalls segments establishing causal dependency, with the consolidated L5 profile connecting her career aspiration to formative experiences. The recall gating module excludes memories lacking true relevance. This structured causality yields the correct answer: \textbf{\textit{No, she likely wouldn't}}.

\textbf{Mem0} as a representative baseline recalls fragmented factual memories. Without hierarchical consolidation, the framework fails to construct the \textit{support$\rightarrow$career chain}, producing an inverted answer: \textbf{\textit{Yes, she would}}.

This comparison highlights how TMT's temporal containment and instruction-based consolidation organize episodic evidence into a coherent inferential structure for counterfactual reasoning.

\vspace{-0.4em}
\subsection{Discussion}
\vspace{-0.4em}

Our experiments suggest three key takeaways for long-horizon memory in conversational agents.

\textbf{Temporal continuity is an effective organizing principle.} By enforcing temporal containment, TMT provides stable temporal leaves for consolidation and recall, instead of treating semantic similarity as the primary structure. Ablation studies and manifold analysis indicate that this temporal hierarchy enables effective compression: it facilitates the construction of temporal evidence chains, amplifies user-specific distinctions, and suppresses noise in long dialogs.

\textbf{Semantic-guided consolidation makes abstraction explicit and portable.} Level-specific prompts encourage distinct consolidation objectives across layers without architecture-specific tuning. Empirically, hierarchical consolidation outperforms the evaluated methods, indicating that progressive transformation over temporally grouped memories is beneficial beyond storing more text.

\textbf{Recall reflects a practical trade-off between precision and efficiency.} The complexity-aware recall planner consistently outperforms fixed recall scopes, while recall gating is most effective when the candidate set contains distractors. For highly complex queries, broader context may outweigh aggressive filtering, and planner errors can expand or shrink the recall scope; in practice, recall budgets can be tuned to different application needs.

Overall, TiMem suggests that combining temporal organization with hierarchical consolidation and adaptive recall yields compact yet grounded long-term memory for conversational agents.

\vspace{-0.4em}
\section{Conclusion}
\vspace{-0.4em}

\label{sec:conclusion}

We introduced \textbf{TiMem}, a temporal--hierarchical memory framework for long-horizon conversational agents, which treats temporal continuity as a first-class organizing principle for long-term memory personalization.
TiMem provides: (i) the \textbf{TMT}, a structure that enforces temporal containment and order; (ii) \textbf{instruction-guided consolidation without fine-tuning} that progressively transforms raw dialog into higher-level patterns and incrementally refined profiles updated monthly; and (iii) \textbf{complexity-aware recall} that plans the recall scope, propagates evidence hierarchically from activated leaves, and applies recall-time gating to retain only query-relevant memories.

Under a consistent evaluation setup, TiMem achieves state-of-the-art accuracy of \textbf{75.30\%} on \textbf{LoCoMo} and \textbf{76.88\%} on \textbf{LongMemEval-S}, while \textbf{reducing recalled context by 52.20\%} on LoCoMo via recall planning and gating.
Manifold analysis indicates that temporal consolidation yields persona separation while reducing dispersion, supporting coherent long-horizon memory representations.

We view TiMem as a practical and interpretable foundation for long-term agent memory. Future directions include combining temporal hierarchies with richer structured memory representations and incorporating storage-time forgetting and adaptive temporal boundaries to further improve efficiency and robustness.

\section{Limitations}
\label{sec:limitations}

\paragraph{LLM Middleware Performance} Consolidation and recall modules rely on general-purpose LLMs through instruction prompts. Fine-tuning specialized smaller models for these operations may improve efficiency while maintaining module functionality.

\paragraph{Structured Representation} High-level memories lack explicit categorical structures or knowledge graphs. Hybrid architectures combining temporal hierarchies with typed entity representations may better capture multi-dimensional content.

\paragraph{Forgetting Mechanism} The framework lacks storage-time forgetting mechanism. Future work should explore effective storage-level forgetting methods that selectively consolidate memories while maintaining critical facts and recurring patterns, balancing efficiency with factual integrity.

\paragraph{Temporal Parameterization} TiMem uses realistic temporal boundaries for reproducibility. Adaptive temporal boundaries detection or interaction-density scheduling could enhance domain transferability.

\section{Ethics Statement}
\label{sec:ethics}
This work does not involve human subjects or personally identifiable information. Experiments use publicly available benchmarks under appropriate licenses. TiMem enforces strict user-group isolation by design: each memory tree is scoped to a single user, with no cross-user memory sharing or aggregation, protecting individual privacy. Deployed systems should implement secure storage, explicit user consent, and data deletion mechanisms. As with any LLM-based system, practitioners should monitor for potential biases in memory consolidation and ensure transparency about retention policies.

\section*{Acknowledgments}
This work was supported in part by the National Science and Technology Major Project of China under Grant 2025ZD1607304, Beijing Natural Science Foundation under Grant L243009,the National Nature Science Foundation of China under Grant 62473367 and the Science and Technology Service Network Initiative (STS) Project of Chinese Academy of Sciences under Grant STS-HP-202308.

\bibliography{references}

\appendix

\section{Dataset Details}
\label{app:datasets}

\subsection{LoCoMo Benchmark}
\textbf{LoCoMo}~\cite{locomo} consists of 10 user groups and multi-session conversations spanning over six months on average. Conversations are grounded in dialog streams with explicit timestamps. We use 1,540 questions in the benchmark covering (1) single-hop, (2) multi-hop, (3) open-domain, and (4) temporal reasoning.

\subsection{LongMemEval-S Benchmark}
\textbf{LongMemEval-S}~\cite{longmemeval} is a synthetic benchmark with 500 conversations and 500 questions. It simulates online memory processing and assesses five capabilities: (1) information extraction from single user/assistant turns, (2) multi-session reasoning, (3) knowledge updates, (4) temporal reasoning, and (5) abstention. The benchmark constructs 500 user personas by sampling attributes from a shared persona template pool.

\section{Implementation Details}
\label{app:implementation}

\subsection{Memory Consolidation Configuration}
All frameworks use \texttt{gpt-4o-mini-2024-07-18} for memory consolidation with temperature 0.7 and \texttt{Qwen3-Embedding-0.6B} (dimension 1024) for embeddings, ensuring consistent processing capabilities across comparisons. For TiMem: L1 memories created online using non-overlapping sliding window with segment size $w_d=1$ turn (one user--assistant exchange); L2--L5 generated through scheduled aggregation at temporal boundaries (session end, daily, weekly, and $\mathcal{G}_5$ monthly intervals for profiles). Historical context window $w_i=3$ prior memories per layer, balancing continuity and computational cost.

\newcommand{\gptmini}{\texttt{gpt-4o-mini-\allowbreak 2024-\allowbreak 07-\allowbreak 18}}
\newcommand{\gptfour}{\texttt{gpt-4o-\allowbreak 2024-\allowbreak 11-\allowbreak 20}}
\newcommand{\qwenemb}{\texttt{Qwen3-Embedding-\allowbreak 0.6B}}

\subsection{Question Answering Configuration}

\noindent\textbf{Basic settings} Baselines are evaluated using their recommended configurations with the following unified settings:
\gptmini{} for consolidation and recall, \qwenemb{} for embeddings, and recall budget $k{=}20$ memories.

\par\noindent\textbf{LoCoMo} Question answering uses \gptmini{} with Mem0's prompt templates.

\par\noindent\textbf{LongMemEval-S} We use a unified template to clearly separate roles.
\emph{Internal LLM} (consolidation/planner/gating) is \gptmini{}.
\emph{External LLM} is \gptmini{} by default (we additionally report results with \gptfour{}).
\emph{Judge LLM} follows the official LongMemEval-S evaluation prompt.
The evaluation prompt was meta-evaluated by its authors to achieve $>$97\% agreement with human experts, supporting the reliability of LLJ-based scoring on this benchmark~\citep{longmemeval}.

\subsection{Recall Configuration}
\label{app:recall_config}

TiMem's hierarchical recall uses complexity-aware configurations with two LLM calls per query: a \textbf{recall planner} (1 call) and \textbf{recall gating} (1 call).

\begin{table}[t]
    \centering
    \small
    \setlength{\tabcolsep}{4pt}
    \begin{tabular}{p{0.22\linewidth}p{0.74\linewidth}}
    \toprule
    \textbf{Stage} & \textbf{Operation and Key Parameters} \\
    \midrule
    \multicolumn{2}{l}{\textbf{1. Recall Planner (1 LLM call)}} \\
    & Predicts complexity $c\in\{\text{simple},\text{hybrid},$ $\text{complex}\}$ and extracts keywords $K$ to set level-specific budgets and search scope $\mathcal{S}(c)$. \\
    \midrule
    \multicolumn{2}{l}{\textbf{2. Hierarchical Recall (no LLM calls)}} \\
    Leaf Activation & Score $L_1$ leaves by $s(m,q,K)=\lambda s_{\text{sem}}+(1-\lambda)s_{\text{lex}}$ with $\lambda{=}0.9$ (cosine similarity + BM25), then select top-$k_1{=}20$. \\
    Ancestor Collection & For each activated leaf, collect ancestors whose levels satisfy $\ell(m)\in\mathcal{S}(c)$ (deterministic traversal). \\
    Budgeting & Keep up to: \textbf{Simple} ($L_1{:}20$, $L_2{:}4$, $L_5{:}1$); \textbf{Hybrid} ($L_1{:}20$, $L_2{:}4$, $L_3{:}2$, $L_5{:}1$); \textbf{Complex} ($L_1{:}20$, $L_2{:}8$, $L_3{:}4$, $L_4{:}2$, $L_5{:}1$). \\
    Pruning / Early Stop & If candidates exceed per-level budgets, prune by similarity scores; if fewer ancestors exist, terminate early. \\
    \midrule
    \multicolumn{2}{l}{\textbf{3. Recall Gating (1 LLM call)}} \\
    & Prompt an LLM to retain/drop each candidate memory conditioned on $(q,c)$, producing the final memory set $\Omega_{\mathrm{final}}$. \\
    \bottomrule
    \end{tabular}
    \caption{Recall configuration in TiMem, organized by the three major stages: recall planner, hierarchical recall, and recall gating.}
    \label{tab:appendix_recall_config}
\end{table}

\section{Parameter Studies}
\label{app:param_studies}

\subsection{LLM Configuration Analysis}

We investigate the interplay between internal LLMs (used for memory consolidation and recall) and external LLMs (used for question answering). This experiment examines the memory system's quality and downstream reasoning capability separately, analyzing how different model combinations affect end-to-end performance.

\paragraph{Experimental Design} We test two internal LLM configurations:
\begin{itemize}[leftmargin=*, itemsep=0.2em, topsep=0.2em, parsep=0pt]
    \item \textbf{GPT-4o-mini}: A representative commercial LLM used for memory consolidation and recall.
    \item \textbf{Qwen3-32B}: A representative open-source LLM used as a production-oriented alternative.
\end{itemize}

As shown in table \ref{tab:param_p1}, for each internal configuration, we evaluate five external LLMs for question answering: \texttt{gpt-4o-mini}, \texttt{gpt-4o}, \texttt{qwen3-8b}, \texttt{qwen3-32b}, and \texttt{qwen3-235b-a22b}. We report results from two LLM-as-judge evaluators: \texttt{gpt-4o-mini} (denoted LLJ-G) and \texttt{qwen3-32b} (denoted LLJ-Q), and observe consistent trends across both judges.

\input{tables/param_p1_llm_config.tex}

\vspace{-0.8em}

\subsection{Segment Granularity Analysis}

We analyze the impact of L1 segment size on memory quality and retrieval effectiveness. Segment granularity determines the trade-off between detail preservation and computational efficiency.

\paragraph{Experimental Design.} We evaluate four segment sizes based on dialog turn counts: 1 turn (finest), 2 turns, 4 turns, and 8 turns (coarsest). All other parameters use default values.

\input{tables/param_p3_segment_size.tex}
\vspace{-0.4em}

As shown in table \ref{tab:param_p3}, the performance of the question-answering decreases as the size of the segment increases, indicating a trade-off between the size of the segment and the accuracy of the QA in practical applications.
\vspace{-0.4em}
\subsection{Sensitivity Analysis of $\lambda$}
\label{app:lambda_sensitivity}

\begin{table}[ht]
\centering
\small
\begin{tabular}{ccc}
\toprule
$\lambda$ & LoCoMo LLJ (\%) $\uparrow$ & $\Delta$ vs. $\lambda=0.9$ \\
\midrule

0.7 & 73.96 & -1.34  \\
0.8 & 74.55 & -0.75  \\
0.9 & 75.30 & --  \\
1.0 & 74.68 & -0.62 \\
\bottomrule
\end{tabular}
\caption{Sensitivity analysis of the semantic--lexical balance parameter $\lambda$ on the LoCoMo dataset.}
\label{tab:lambda_sensitivity}
\end{table}

To validate the robustness of the hybrid retrieval mechanism, we conduct a sensitivity sweep for $\lambda$ on LoCoMo, experiment setup is consistent with Chapter \ref{sec:experiments}. Table~\ref{tab:lambda_sensitivity} shows that performance remains consistently high across values, with a mild peak at $\lambda=0.9$ The stability across these configurations demonstrates that the framework effectively integrates dual-channel evidence without overfitting to specific benchmark distributions.

\section{Memory Manifold Analysis}
\label{app:manifold}

We analyze how hierarchical consolidation transforms memory structure using manifold metrics: Intrinsic Dimensionality, Silhouette Score, Spread, and Trustworthiness.

\subsection{LoCoMo: Feature Distillation}

Table~\ref{tab:manifold_locomo} shows progressive user differentiation across hierarchy levels. L1 segments exhibit high dimensionality and low clustering quality, indicating that generic conversational patterns dominate at the segment level. Through hierarchical consolidation, dimensionality compresses by 5.6-fold to reach 13 dimensions at L5, while clustering quality improves by 6.2-fold to achieve 0.574 silhouette score. The separation ratio increases from 0.30 to 2.14, indicating extraction of user-specific features from dialog streams.

\input{tables/analysis_manifold_locomo.tex}

\subsection{LongMemEval-S: Noise Suppression}

Table~\ref{tab:manifold_lmes} reveals convergence toward shared structure in the synthetic dataset. L1 exhibits high spread at 0.692 and maximum dimensionality at 100. Through consolidation, spread reduces by 50\% to reach 0.345 at L5, while the effective radius (mean distance to centroid) shrinks from 0.789 to 0.444. Dimensionality remains saturated at 100 through L1-L4, then drops to 68 at L5, with the low-dimensional shared structure emerging through progressive consolidation.

\input{tables/analysis_manifold_lmes.tex}

\subsection{Adaptive Consolidation}

TiMem demonstrates adaptive consolidation that responds to different data characteristics. In LoCoMo conversations, the framework acts as a feature separator, increasing inter-user variance as separation ratio grows from 0.30 to 2.14. In synthetic LongMemEval-S data, it functions as a noise filter, reducing variance as spread decreases from 0.692 to 0.345. Both processes achieve semantic compression through dimensionality reduction, yet produce contrasting topological effects: expanding distinctiveness for diverse users versus contracting dispersion for noisy data. Trustworthiness scores exceeding 0.78 across all levels indicate that these manifold transformations preserve neighborhood relationships during dimensionality reduction.

\input{prompts/prompts.tex}

\end{document}

%% file: tables/result_r1_locomo.tex
\begin{table*}[!t]
\centering
\scriptsize
\resizebox{\textwidth}{!}{%
\begin{tabular}{l|cccc|ccc}
\toprule
	\textbf{Method} & \textbf{Single-Hop} & \textbf{Temporal} & \textbf{Open-Domain} & \textbf{Multi-Hop} & \textbf{Overall} & \textbf{F1} & \textbf{RL} \\
 & $\uparrow$ \textbf{LLJ} (841Q) & $\uparrow$ \textbf{LLJ} (321Q) & $\uparrow$ \textbf{LLJ} (96Q) & $\uparrow$ \textbf{LLJ} (282Q) & $\uparrow$ \textbf{LLJ} (1540Q) & $\uparrow$ & $\uparrow$ \\
\midrule
MemoryBank & 46.18 $\pm$ 0.32 & 29.34 $\pm$ 0.45 & 36.67 $\pm$ 0.47 & 33.36 $\pm$ 0.54 & 39.77 $\pm$ 0.27 & 25.78 & 25.15 \\
A-MEM & 52.82 $\pm$ 0.28 & 60.87 $\pm$ 0.37 & 43.75 $\pm$ 0.00 & 38.37 $\pm$ 0.27 & 51.29 $\pm$ 0.06 & 30.37 & 36.85 \\
Mem0 & 62.09 $\pm$ 0.42 & 59.25 $\pm$ 0.41 & 37.70 $\pm$ 0.47 & 50.14 $\pm$ 0.32 & 57.79 $\pm$ 0.34 & 42.52 & 44.14 \\
MemoryOS & 68.37 $\pm$ 0.46 & 52.46 $\pm$ 0.49 & \underline{46.67 $\pm$ 1.79} & 52.76 $\pm$ 0.49 & 60.79 $\pm$ 0.48 & \underline{45.36} & 43.74 \\
MemOS & \underline{76.07 $\pm$ 0.10} & \underline{69.47 $\pm$ 0.25} & 45.14 $\pm$ 0.49 & \underline{56.85 $\pm$ 0.69} & \underline{69.24 $\pm$ 0.11} & 45.02 & \underline{47.41} \\
\midrule
\rowcolor{gray!10} \textbf{TiMem (Ours)} & \textbf{81.43 $\pm$ 0.05} & \textbf{77.63 $\pm$ 0.34} & \textbf{52.08 $\pm$ 0.74} & \textbf{62.20 $\pm$ 0.82} & \textbf{75.30 $\pm$ 0.16} & \textbf{54.40} & \textbf{54.68} \\
\bottomrule
\end{tabular}
}
\caption{Performance comparison on LoCoMo benchmark. Categories include Single-Hop, Temporal, Open-Domain, and Multi-Hop. Best results are \textbf{bolded}; \underline{underline} indicates second best.}
\label{tab:locomo_results}
\end{table*}

%% file: tables/result_r2_longmemeval.tex
\begin{table*}[!t]
\centering
\small
\renewcommand{\arraystretch}{1.05}
\resizebox{\textwidth}{!}{%
\begin{tabular}{l|cccccc|c}
\toprule
\multirow{2}{*}{\textbf{Method}} & \multicolumn{6}{c|}{\textbf{LongMemEval-S Task Categories}} & \multirow{2}{*}{\textbf{Overall}} \\
& \textbf{KU} & \textbf{MS} & \textbf{SSA} & \textbf{SSP} & \textbf{SSU} & \textbf{TR} & \\
& $\uparrow$ (78Q) & $\uparrow$ (133Q) & $\uparrow$ (56Q) & $\uparrow$ (30Q) & $\uparrow$ (70Q) & $\uparrow$ (133Q) & $\uparrow$ (500Q) \\
\midrule
\multicolumn{8}{c}{\textit{Answer Model: GPT-4o-mini-2024-07-18}} \\
\midrule
MemoryBank & 21.79 $\pm$ 0.00 & 9.77 $\pm$ 0.00 & 50.00 $\pm$ 0.00 & 12.00 $\pm$ 1.83 & 29.71 $\pm$ 0.64 & 17.14 $\pm$ 0.34 & 21.04 $\pm$ 0.09 \\
A-MEM & 72.82 $\pm$ 0.51 & 40.30 $\pm$ 0.37 & \textbf{87.50 $\pm$ 0.00} & 39.33 $\pm$ 2.49 & 82.86 $\pm$ 0.00 & 36.09 $\pm$ 0.48 & 55.44 $\pm$ 0.15 \\
Mem0 & \underline{78.72 $\pm$ 0.70} & \underline{66.17 $\pm$ 0.92} & 51.79 $\pm$ 0.00 & 50.00 $\pm$ 2.36 & \underline{94.29 $\pm$ 0.00} & 49.17 $\pm$ 0.67 & 64.96 $\pm$ 0.41 \\
MemoryOS & 56.15 $\pm$ 0.57 & 44.81 $\pm$ 0.41 & 78.18 $\pm$ 0.00 & \underline{51.33 $\pm$ 1.83} & 81.14 $\pm$ 0.64 & 53.38 $\pm$ 0.00 & 58.04 $\pm$ 0.18 \\
MemOS & 76.67 $\pm$ 0.51 & 58.80 $\pm$ 0.30 & 67.86 $\pm$ 0.00 & 50.67 $\pm$ 1.33 & 93.71 $\pm$ 0.70 & \underline{65.11 $\pm$ 0.37} & \underline{68.68 $\pm$ 0.16} \\
\midrule
\rowcolor{gray!10} \textbf{TiMem (Ours)} & \textbf{86.16 $\pm$ 1.07} & \textbf{70.83 $\pm$ 0.98} & \underline{82.14 $\pm$ 0.00} & \textbf{63.33 $\pm$ 0.00} & \textbf{95.71 $\pm$ 0.00} & \textbf{68.42 $\pm$ 0.00} & \textbf{76.88 $\pm$ 0.30} \\
\midrule
\multicolumn{8}{c}{\textit{Answer Model: GPT-4o-2024-11-20}} \\
\midrule
MemoryBank & 22.56 $\pm$ 0.70 & 12.78 $\pm$ 0.00 & 61.43 $\pm$ 0.98 & 13.33 $\pm$ 0.00 & 33.43 $\pm$ 0.78 & 13.53 $\pm$ 0.00 & 22.88 $\pm$ 0.23 \\
A-MEM & \underline{87.18 $\pm$ 0.00} & 45.26 $\pm$ 0.30 & \underline{83.21 $\pm$ 0.87} & 56.67 $\pm$ 2.98 & 90.00 $\pm$ 0.00 & 46.77 $\pm$ 0.30 & 63.40 $\pm$ 0.33 \\
Mem0 & 84.87 $\pm$ 0.57 & 65.11 $\pm$ 0.41 & 55.00 $\pm$ 0.80 & \underline{60.67 $\pm$ 1.49} & \underline{95.71 $\pm$ 0.00} & 51.88 $\pm$ 0.00 & 67.56 $\pm$ 0.30 \\
MemoryOS & 60.00 $\pm$ 0.57 & 51.13 $\pm$ 0.53 & 80.00 $\pm$ 0.00 & 53.33 $\pm$ 0.00 & 82.86 $\pm$ 0.00 & 54.59 $\pm$ 0.67 & 61.20 $\pm$ 0.23 \\
MemOS & 76.07 $\pm$ 0.60 & \underline{68.42 $\pm$ 0.00} & 63.69 $\pm$ 0.84 & \textbf{64.44 $\pm$ 1.57} & 92.86 $\pm$ 0.00 & \underline{71.43 $\pm$ 0.61} & \underline{73.07 $\pm$ 0.25} \\
\midrule
\rowcolor{gray!10} \textbf{TiMem (Ours)} & \textbf{87.69 $\pm$ 0.70} & \textbf{72.78 $\pm$ 0.34} & \textbf{85.71 $\pm$ 0.00} & 55.33 $\pm$ 1.83 & \textbf{96.28 $\pm$ 0.78} & \textbf{73.38 $\pm$ 1.14} & \textbf{78.96 $\pm$ 0.26} \\
\bottomrule
\end{tabular}%
}
\caption{Performance comparison on the LongMemEval-S benchmark, reporting LLJ accuracy by task type. Categories include KU: Knowledge Update, MS: Multi-Session, SSA/P/U: Single-Session Assistant/Preference/User, and TR: Temporal Reasoning. Best results are \textbf{bolded}; \underline{underline} indicates second best.}

\label{tab:longmemeval_detailed}
\end{table*}

%% file: tables/ablation_e1_routing_filter.tex
\begin{table}[!t]
\centering
\resizebox{\columnwidth}{!}{%
\begin{tabular}{l|cc|cc}
\toprule
	\multirow{2}{*}{\textbf{Configuration}} & \multicolumn{2}{c|}{\textbf{LoCoMo}} & \multicolumn{2}{c}{\textbf{LongMemEval-S}} \\
	& \textbf{LLJ} $\uparrow$ & \textbf{Mem Len} $\downarrow$ & \textbf{LLJ} $\uparrow$ & \textbf{Mem Len} $\downarrow$ \\
\midrule
\multicolumn{5}{l}{\cellcolor{gray!10}\textit{w/o Planner, w/o Gating}} \\
\quad - Simple & 73.51 & 3710.30 & 73.20 & 3371.53 \\
\quad - Hybrid & 72.40 & 4376.40 & 74.00 & 4054.78 \\
\quad - Complex & 72.86 & 5658.26 & 74.40 & 5685.68 \\
\midrule
\multicolumn{5}{l}{\cellcolor{gray!10}\textit{w/o Planner, w Gating}} \\
\quad - Simple & 71.88 & \textbf{367.68} & 69.00 & \textbf{397.04} \\
\quad - Hybrid & \underline{73.38} & 691.59 & \underline{75.00} & 1673.93 \\
\quad - Complex & 72.92 & 4479.06 & 74.20 & 3028.68 \\
\midrule
\multicolumn{5}{l}{\cellcolor{gray!10}\textit{w Planner, w/o Gating}} \\
\quad - Planned & 72.99 & 4411.09 & 73.80 & 3941.98 \\
\midrule
\textbf{w P., w G. (Baseline)} & \textbf{75.30} & \underline{511.25} & \textbf{76.88} & \underline{1270.62} \\
\bottomrule
\end{tabular}%
}
\caption{Recall Planner and Recall Gating effectiveness. The planned configuration of the TiMem baseline achieves the best balance between accuracy and memory length.}
\label{tab:ablation_e1}
\end{table}

%% file: tables/ablation_e2_hierarchy.tex
\begin{table}[!t]
\centering
\resizebox{\columnwidth}{!}{%
\begin{tabular}{l|cc|cc}
\toprule
	\multirow{2}{*}{\textbf{Memory Layers}} & \multicolumn{2}{c|}{\textbf{LoCoMo}} & \multicolumn{2}{c}{\textbf{LongMemEval-S}} \\
	& \textbf{LLJ} $\uparrow$ & \textbf{Mem Len} $\downarrow$ & \textbf{LLJ} $\uparrow$ & \textbf{Mem Len} $\downarrow$ \\
\midrule
\multicolumn{5}{l}{\cellcolor{gray!10}\textit{L1 only (base layer)}} \\
\quad w Flat Rec. & 70.06 & 995.15 & 57.40 & 1823.98 \\
\quad w Hier. Rec. & \underline{73.18} & \textbf{361.23} & \underline{72.40} & \textbf{437.42} \\
\midrule
\multicolumn{5}{l}{\cellcolor{gray!10}\textit{L2-L5 only (high-level)}} \\
\quad w Flat Rec. & 51.23 & 2348.49 & 48.00 & 2657.68 \\
\quad w Hier. Rec. & 57.08 & 3786.44 & 64.20 & 2344.92 \\
\midrule
\multicolumn{5}{l}{\cellcolor{gray!10}\textit{L1-L5 (full hierarchy)}} \\
\quad w Flat Rec. & 70.71 & 1715.65 & 55.40 & 4519.26 \\
	\rowcolor{gray!10} \quad \textbf{w H. R. (Baseline)} & \textbf{75.30} & \underline{511.25} & \textbf{76.88} & \underline{1270.62} \\
\bottomrule
\end{tabular}%
}
\caption{ Hierarchical vs. Flat Architectures. Comparison of L1-only, L2-L5 only, and Full Hierarchy with flat vs. hierarchical recall strategies.}
\label{tab:ablation_e2}
\end{table}

%% file: tables/analysis_efficiency_metrics.tex
\begin{table}[!ht]
\centering
\small
\resizebox{\columnwidth}{!}{%
\begin{tabular}{l|cc|cc}
\toprule
\textbf{Method} & \multicolumn{2}{c|}{\textbf{LoCoMo}} & \multicolumn{2}{c}{\textbf{LongMemEval-S}} \\
& \textbf{Memory} & \textbf{Latency} & \textbf{Memory} & \textbf{Latency} \\
& $\downarrow$ (tokens) & $\downarrow$ (P50/P95) & $\downarrow$ (tokens) & $\downarrow$ (P50/P95) \\
\midrule
MemoryBank & 8063.77 & 9.46/13.07 & 13906.81 & 10.74/14.50 \\
A-MEM & 2431.4 & 1.74/7.23 & 3971.6 & 5.12/11.89 \\
Mem0 & \underline{1070.10} & 2.44/4.29 & 1647.56 & 3.64/6.11 \\
MemoryOS & 4659.09 & \textbf{1.66/2.21} & 7574.30 & \textbf{1.63}/\underline{3.95} \\
MemOS & 1371.42 & \underline{1.69/3.44} & \textbf{1091.51} & \underline{1.64}/\textbf{2.70} \\
\midrule
\rowcolor{gray!10} TiMem (Ours) & \textbf{511.25} & 2.35/4.91 & \underline{1270.62} & 1.76/4.48 \\
\bottomrule
\end{tabular}%
}
\caption{\textbf{Recall efficiency metrics.} Memory context length and P50/P95 latency across benchmarks. TiMem significantly reduces context load compared to baselines while maintaining low latency.}
\label{tab:efficiency_metrics}
\end{table}

%% file: tables/param_p1_llm_config.tex

\begin{table}[!ht]
\centering
\small
\resizebox{\columnwidth}{!}{%
\begin{tabular}{@{}lcccc@{}}
\toprule
\textbf{Answer Model} & \textbf{F1} $\uparrow$ & \textbf{RL} $\uparrow$ & \textbf{LLJ-G} $\uparrow$ & \textbf{LLJ-Q} $\uparrow$ \\
\midrule
\multicolumn{5}{c}{\cellcolor{gray!10}\textit{Internal: GPT-4o-mini-2024-07-18}} \\
\midrule
GPT-4o-mini        & \underline{54.40} & \underline{54.68} & 75.30  & 72.86    \\
GPT-4o             & \textbf{56.14} & \textbf{56.40} & 74.03 & 72.60    \\
Qwen3-8B          & 46.58 & 47.00 & 66.04 & 64.68    \\
Qwen3-32B          & 50.45 & 50.82 & \underline{74.61} & \underline{73.38}    \\
Qwen3-235B-A22B    & 42.17 & 43.60 & \textbf{80.45} & \textbf{79.03}    \\
\midrule
\multicolumn{5}{c}{\cellcolor{gray!10}\textit{Internal: Qwen3-32B}} \\
\midrule
GPT-4o-mini        & \underline{51.50} & \underline{51.74} & 71.23 & 70.00    \\
GPT-4o             & \textbf{53.65} & \textbf{53.90} & 72.60 & 70.19    \\
Qwen3-8B          & 44.59 & 45.29 & 64.16 & 62.34    \\
Qwen3-32B          & 46.74 & 47.15 & \underline{74.74} & \underline{72.73} \\
Qwen3-235B-A22B    & 40.20 & 41.47 & \textbf{77.73} & \textbf{76.23}    \\
\bottomrule
\end{tabular}%
}
\caption{LLM Configuration Analysis. Comparison of answering models across two internal memory generation models (GPT-4o-mini, Qwen3-32B).}
\label{tab:param_p1}
\end{table}

%% file: tables/param_p3_segment_size.tex

\begin{table}[!ht]
\centering
\footnotesize
\begin{tabular*}{\columnwidth}{@{\extracolsep{\fill}}lcccc}
\toprule
\textbf{Segment Size} & \textbf{F1} & \textbf{RL} & \textbf{LLJ (\%)} & \textbf{Delta} \\
\midrule
1 turn (baseline)                & 54.40    & 54.68    & 75.30    & -- \\
2 turns                        & 52.45    & 51.00    & 73.64    & -1.66 \\
4 turns              & 50.56    & 49.27    & 70.00    & -5.30 \\
8 turns           & 46.94    & 45.72    & 65.26    & -10.04 \\
\bottomrule
\end{tabular*}
\caption{Impact of L1 segment granularity on LoCoMo performance. Segment size determines the number of dialogue turns aggregated into each L1 memory. Default configuration uses 1 turn. Delta shows performance change relative to 1-turn baseline.}
\label{tab:param_p3}
\end{table}

%% file: tables/analysis_manifold_locomo.tex
\begin{table}[t]
\centering
\small
\resizebox{\columnwidth}{!}{%
\begin{tabular}{lccccc}
\toprule
\textbf{Layer} & \textbf{IntDim}$\downarrow$ & \textbf{Silh}$\uparrow$ & \textbf{Sep.Ratio}$\uparrow$ & \textbf{Trust}$\uparrow$ & \textbf{Cont}$\uparrow$ \\
\midrule
L1 & 73 & 0.093 & 0.30 & 0.950 & 0.941 \\
L2 & 35 & 0.273 & 0.77 & 0.964 & 0.955 \\
L3 & 37 & 0.274 & 0.76 & 0.963 & 0.955 \\
L4 & 28 & 0.329 & 0.89 & 0.972 & 0.964 \\
L5 & 13 & \textbf{0.574} & \textbf{2.14} & 0.818 & 0.862 \\
\bottomrule
\end{tabular}%
}
\caption{\textbf{LoCoMo manifold metrics.} Progressive feature separation from L1 to L5 evidenced by increasing Silhouette Score and Separation Ratio.}
\label{tab:manifold_locomo}
\end{table}

%% file: tables/analysis_manifold_lmes.tex
\begin{table}[t]
\centering
\small
\resizebox{\columnwidth}{!}{%
\begin{tabular}{lccccc}
\toprule
\textbf{Layer} & \textbf{IntDim}$\downarrow$ & \textbf{Spread}$\downarrow$ & \textbf{CV}$\downarrow$ & \textbf{Radius95}$\downarrow$ & \textbf{Trust}$\uparrow$ \\
\midrule
L1 & 100 & 0.692 & 0.085 & 0.789 & 0.917 \\
L2 & 100 & 0.672 & 0.078 & 0.761 & 0.942 \\
L3 & 100 & 0.533 & 0.162 & 0.669 & 0.847 \\
L4 & 82 & 0.432 & 0.180 & 0.575 & 0.812 \\
L5 & 68 & \textbf{0.345} & 0.155 & \textbf{0.444} & 0.789 \\
\bottomrule
\end{tabular}%
}
\caption{\textbf{LongMemEval-S convergence metrics.} Reduced Spread and Radius95 indicate convergence toward unified persona templates from L1 to L5.}
\label{tab:manifold_lmes}
\end{table}

%% file: prompts/prompts.tex
\onecolumn
\section{Prompt Templates}
\label{app:prompt_template}

\definecolor{promptbg}{RGB}{248,249,250}
\definecolor{promptframe}{RGB}{108,117,125}
\definecolor{prompttitlebg}{RGB}{52,103,172}
\definecolor{prompttitlefg}{RGB}{255,255,255}

\newtcolorbox{promptbox}[1][]{%
  enhanced,
  breakable,
  colback=promptbg,
  colframe=promptframe!50,
  coltitle=prompttitlefg,
  colbacktitle=prompttitlebg,
  fonttitle=\small\bfseries\ttfamily,
  title={#1},
  boxrule=0.5pt,
  arc=4pt,
  left=8pt,
  right=8pt,
  top=6pt,
  bottom=6pt,
  toptitle=3pt,
  bottomtitle=3pt,
  before skip=2pt,
  after skip=6pt,
  pad at break*=2pt,
  title after break={\small\itshape (continued)},
  bottomrule at break={0.5pt},
  toprule at break={0.5pt},
}

\newtcolorbox{promptboxplain}{%
  enhanced,
  breakable,
  colback=promptbg,
  colframe=promptframe!50,
  boxrule=0.5pt,
  arc=4pt,
  left=8pt,
  right=8pt,
  top=6pt,
  bottom=6pt,
  before skip=2pt,
  after skip=6pt,
  pad at break*=2pt,
  bottomrule at break={0.5pt},
  toprule at break={0.5pt},
}

\newcommand{\promptbody}{\small\ttfamily\raggedright}

\subsection{LoCoMo Benchmark}
\label{app:prompt_locomo}

We adopt the prompt template from Mem0~\cite{mem0} for LoCoMo QA and evaluation:

\subsubsection{Question Answering Prompt}
\label{app:prompt_locomo_qa}
\noindent
\begin{promptbox}[LoCoMo QA Prompt]
\promptbody
You are an intelligent memory assistant tasked with retrieving accurate information from conversation memories.

\textbf{\# CONTEXT:}\\
You have access to memories from two speakers in a conversation. These memories contain timestamped information that may be relevant to answering the question.

\textbf{\# INSTRUCTIONS:}\\
1. Carefully analyze all provided memories from both speakers\\
2. Pay special attention to the timestamps to determine the answer\\
3. If the question asks about a specific event or fact, look for direct evidence in the memories\\
4. If the memories contain contradictory information, prioritize the most recent memory\\
5. If there is a question about time references (like "last year", "two months ago", etc.), calculate the actual date based on the memory timestamp. For example, if a memory from 4 May 2022 mentions "went to India last year," then the trip occurred in 2021.\\
6. Always convert relative time references to specific dates, months, or years. For example, convert "last year" to "2022" or "two months ago" to "March 2023" based on the memory timestamp. Ignore the reference while answering the question.\\
7. Focus only on the content of the memories from both speakers. Do not confuse character names mentioned in memories with the actual users who created those memories.\\
8. The answer should be less than 5-6 words.

\textbf{\# APPROACH (Think step by step):}\\
1. First, examine all memories that contain information related to the question\\
2. Examine the timestamps and content of these memories carefully\\
3. Look for explicit mentions of dates, times, locations, or events that answer the question\\
4. If the answer requires calculation (e.g., converting relative time references), show your work\\
5. Formulate a precise, concise answer based solely on the evidence in the memories\\
6. Double-check that your answer directly addresses the question asked\\
7. Ensure your final answer is specific and avoids vague time references

Relevant Memories:\\
\{context\_memories\}

Question: \{question\}

Answer:
\end{promptbox}

\subsubsection{LLM-as-Judge Evaluation Prompt}
\label{app:prompt_locomo_judge}
\noindent
\begin{promptbox}[LoCoMo LLM-as-Judge]
\promptbody
Your task is to label an answer to a question as 'CORRECT' or 'WRONG'. You will be given the following data:\\
\hspace*{1em}(1) a question (posed by one user to another user),\\
\hspace*{1em}(2) a 'gold' (ground truth) answer,\\
\hspace*{1em}(3) a generated answer\\
which you will score as CORRECT/WRONG.

The point of the question is to ask about something one user should know about the other user based on their prior conversations. The gold answer will usually be a concise and short answer that includes the referenced topic, for example:\\
Question: Do you remember what I got the last time I went to Hawaii?\\
Gold answer: A shell necklace\\
The generated answer might be much longer, but you should be generous with your grading - as long as it touches on the same topic as the gold answer, it should be counted as CORRECT.

For time related questions, the gold answer will be a specific date, month, year, etc. The generated answer might be much longer or use relative time references (like "last Tuesday" or "next month"), but you should be generous with your grading - as long as it refers to the same date or time period as the gold answer, it should be counted as CORRECT. Even if the format differs (e.g., "May 7th" vs "7 May"), consider it CORRECT if it's the same date.

Now it's time for the real question:\\
Question: \{question\}\\
Gold answer: \{standard\_answer\}\\
Generated answer: \{generated\_answer\}

First, provide a short (one sentence) explanation of your reasoning, then finish with CORRECT or WRONG. Do NOT include both CORRECT and WRONG in your response, or it will break the evaluation script.

Just return the label CORRECT or WRONG in a json format with the key as "label".
\end{promptbox}

\subsection{LongMemEval-S Benchmark}
\label{app:prompt_lme}

\subsubsection{Question Answering Prompt}
\label{app:prompt_lme_qa}
We follow the default non-CoT template from LongMemEval~\cite{longmemeval}:
\noindent
\begin{promptbox}[LongMemEval QA Prompt]
\promptbody
I will give you several related memories between you and a user. Please answer the question based on the relevant memories.

Related Memories:

\{memories\}

Current Date: \{current\_date\}\\
Question: \{question\}\\
Answer:
\end{promptbox}

\subsubsection{LLM-as-Judge Evaluation Prompt}
\label{app:prompt_lme_judge}
LongMemEval uses task-specific evaluation prompts. For most tasks (SSU, SSA, MS):
\noindent
\begin{promptbox}[LongMemEval Judge (Standard)]
\promptbody
I will give you a question, a correct answer, and a response from a model. Please answer yes if the response contains the correct answer. Otherwise, answer no. If the response is equivalent to the correct answer or contains all the intermediate steps to get the correct answer, you should also answer yes. If the response only contains a subset of the information required by the answer, answer no.

Question: \{question\}

Correct Answer: \{answer\}

Model Response: \{response\}

Is the model response correct? Answer yes or no only.
\end{promptbox}

For temporal reasoning tasks, off-by-one tolerance is applied:
\noindent
\begin{promptbox}[LongMemEval Judge (Temporal)]
\promptbody
I will give you a question, a correct answer, and a response from a model. Please answer yes if the response contains the correct answer. Otherwise, answer no. If the response is equivalent to the correct answer or contains all the intermediate steps to get the correct answer, you should also answer yes. If the response only contains a subset of the information required by the answer, answer no. In addition, do not penalize off-by-one errors for the number of days. If the question asks for the number of days/weeks/months, etc., and the model makes off-by-one errors (e.g., predicting 19 days when the answer is 18), the model's response is still correct.

Question: \{question\}

Correct Answer: \{answer\}

Model Response: \{response\}

Is the model response correct? Answer yes or no only.
\end{promptbox}

For knowledge update tasks:
\noindent
\begin{promptbox}[LongMemEval Judge (Knowledge Update)]
\promptbody
I will give you a question, a correct answer, and a response from a model. Please answer yes if the response contains the correct answer. Otherwise, answer no. If the response contains some previous information along with an updated answer, the response should be considered as correct as long as the updated answer is the required answer.

Question: \{question\}

Correct Answer: \{answer\}

Model Response: \{response\}

Is the model response correct? Answer yes or no only.
\end{promptbox}

For single-session preference tasks:
\noindent
\begin{promptbox}[LongMemEval Judge (Preference)]
\promptbody
I will give you a question, a rubric for desired personalized response, and a response from a model. Please answer yes if the response satisfies the desired response. Otherwise, answer no. The model does not need to reflect all the points in the rubric. The response is correct as long as it recalls and utilizes the user's personal information correctly.

Question: \{question\}

Rubric: \{rubric\}

Model Response: \{response\}

Is the model response correct? Answer yes or no only.
\end{promptbox}

\subsection{TiMem System Prompts}
\label{app:prompt_timem}

We present key prompt templates used in TiMem's internal processing pipeline:

\subsubsection{L1 Segment Memory Consolidator}
\label{app:prompt_timem_consolidator_l1}
\noindent
\begin{promptbox}[L1 Segment Consolidator]
\promptbody
You are a dialogue memory generator. Your task is to write a fragment memory that captures only the NEW facts from the "Current Conversation" (do not repeat anything already covered in "Historical Memories").

\textbf{Core principle:}\\
Convert dialogue from first-person to third-person narration, preserving as much substantive information content from the original as possible, excluding only confirmed non-informative words.

\textbf{What to preserve:}\\
- All substantive information: people, events, times, places, causes, results, numbers, specific descriptions\\
- Original wording: Keep specific terms used in dialogue for titles, item names, activity descriptions, etc, numbers use Arabic numerals\\
- Emotional expressions: Retain explicit emotions and attitudes from original (like "happy", "worried", "likes"), but avoid adding subjective inferences not present in original

\textbf{What to exclude:}\\
Only exclude purely functional words: greetings ("hi""bye"), confirmation words ("uh-huh""okay""yes"), meaningless fillers ("um""you know""like")

\textbf{Time normalization:}\\
- Preserve the original relative time expressions exactly as written (e.g., "last night", "this morning", "last Friday"). DO NOT convert relative time to absolute dates.

\textbf{Style:}\\
- Use English third-person narration.\\
- Write plain sentences (no lists/numbering/Markdown). Aim for 2-4 sentences, but allow longer to retain essential details.\\
- Use exact proper nouns as in the dialogue; do not replace/expand/infer names, organizations, or locations.\\
- Each memory should focus on one core fact or closely related fact group; avoid packing too many unrelated details into a single entry.

\textbf{Inputs:}\\
- Historical memories (do not repeat): \{previous\_summary\}\\
- Current conversation: \{new\_dialogue\}

Please generate a fragment memory that contains ONLY the new facts. If the current conversation has no substantial new content, provide a minimal 1-2 sentence summary of the core topic or attitude expressed in this turn (do NOT output "no significant additions" or similar empty statements).
\end{promptbox}

\subsubsection{L2 Session Memory Consolidator}
\label{app:prompt_timem_consolidator_l2}
\noindent
\begin{promptbox}[L2 Session Consolidator]
\promptbody
You are a session-level memory generator. Your task is to integrate the fragment memories from this session into a factual session-level summary for downstream daily/weekly aggregation. Aim to maximize fact density without losing essentials and minimize redundancy.

\textbf{Core principle:}\\
Integrate into a coherent session summary while preserving all substantive information from fragments. Prioritize detail retention, moderate deduplication to enhance density.

\textbf{Content requirements:}\\
- Present in chronological order: who-did-what-when-where-why/so-what, plus key decisions and state changes\\
- Strictly base on fragment memories; DO NOT invent facts or "reasonable expansions" beyond fragments\\
- Preserve specific information from fragments: numbers, quantifiers, modifiers, relationship terms, parallel specific items, descriptors, using original wording from fragments\\
- Deduplication: Remove greetings and repeated confirmations, merge near-duplicates, keep most complete version when multiple fragments mention same fact

\textbf{Time handling:}\\
- Strictly preserve relative time expressions from fragment memories (e.g., "last night", "this morning", "last week"). DO NOT convert them to absolute dates.\\
- ONLY preserve both relative and absolute time when fragment memories already contain both; NEVER infer/calculate absolute dates from timestamps.

\textbf{Style and length:}\\
- Use English third-person narration. Use natural, coherent narrative sentences; do not use lists, numbering, or any format markers.\\
- Target about 200-300 words; allow modest overflow to preserve essentials, but avoid verbosity and vacuity.\\
- Keep proper nouns exactly as in the dialogue; do not invent/expand names, organizations, or locations.

\textbf{Input:}\\
- Session fragment memories: \{fragment\_summaries\}

Please generate session-level summary.
\end{promptbox}

\subsubsection{L3 Daily Memory Consolidator}
\label{app:prompt_timem_consolidator_l3}
\noindent
\begin{promptbox}[L3 Daily Consolidator]
\promptbody
You are the agent's long-term memory builder. Task: Generate precise event records for today to serve as foundation for memory retrieval and character understanding.

\textbf{Core Positioning:}\\
L3 memory is the factual layer, accurately recording "who did what and said what when" to provide reliable basis for higher-level pattern recognition and profiling.

\textbf{Recording Principles:}\\
- Precision: Preserve specific names of times, people, places, and items\\
- Completeness: Record cause-effect of events and decision rationale\\
- Authenticity: Prioritize direct quotes\\
- Uniqueness: Record details that reveal individual characteristics

\textbf{Must Include:}\\
- Time (specific or relative)\\
- People (use names from dialogue)\\
- Action (specific verb + specific object)\\
- Quotes (at least one per event: "XXX said '...'")\\
- Details (numbers, locations, item names)\\
- Decisions (if any, record options and rationale)

\textbf{Evidence-first and Less-is-more:}\\
- State only well-supported facts; if evidence is weak, omit or keep to one factual sentence\\
- Do not hypothesize or expand beyond provided content\\
- If no direct quote and facts are incomplete, prefer not to write the event

\textbf{Time Expression:} Preserve relative time from dialogue ("yesterday", "last Friday"), do not convert to absolute dates.

\textbf{Output Standards:}\\
- English third-person, 200-400 words\\
- Chronological order, natural paragraphs\\
- 1-2 sentences per event

\textbf{Input:}\\
- Historical daily reports: \{previous\_daily\_memories\}\\
- Current date: \{date\}\\
- Today's sessions: \{session\_summaries\}

Generate today's event records directly:
\end{promptbox}

\subsubsection{L4 Weekly Memory Consolidator}
\label{app:prompt_timem_consolidator_l4}
\noindent
\begin{promptbox}[L4 Weekly Consolidator]
\promptbody
You are the agent's behavioral pattern analyst. Task: Identify individual behavioral patterns and changes from this week's daily reports to help the agent understand character uniqueness.

\textbf{Core Positioning:}\\
L4 memory is the pattern layer, extracting unique practices and preferences through identifying recurring behaviors and trend changes, serving the agent's deep understanding of characters.

\textbf{Analysis Principles:}\\
- Recurring behavior identification: Find similar choices or practices appearing multiple times this week\\
- Uniqueness excavation: Explain how this person differs from typical people\\
- Change tracking: Mark first-time behaviors or frequency changes\\
- Data support: Support analysis with specific counts, times, and cases\\
- Less-is-more: If evidence is insufficient, skip extracting the pattern; do not speculate

\textbf{Content Structure:}\\

1. This Week's Event Overview (timeline, preserve dates and key information)

2. Behavioral Pattern Recognition (focus)\\
If recurring behaviors found, record:\\
- Pattern description: Use behaviors not adjectives\\
- Specific cases: List each occurrence this week (time + specific action)\\
- Unique aspects: How this person's approach differs from typical people\\
- Frequency comparison: X times this week vs Y times last week (if data available)\\
- Evidence threshold: If you cannot provide at least 2 cases or explicit frequency this week, skip the pattern

3. Preferences and Tendencies\\
Identified from repeated choices:\\
- Preference content: What repeatedly chosen among options\\
- Choice rationale: Reasons extracted from dialogue\\
- Evidence threshold: If rationale relies on guesswork or lacks quotes, omit the item

4. This Week's Changes\\
- First-time behaviors or topics\\
- Significant frequency increases/decreases\\
- Shifts in approach or attitude\\
- Specific description: From what to what, when started\\
- Evidence threshold: If no clear contrast or time point, do not conclude

\textbf{Wording Standards:}\\
Use behavioral descriptions instead of abstract adjectives. For example:\\
- "consults 2-3 people before each decision" not "is cautious"\\
- "mentioned family 4 times this week" not "values family"\\
- "chose A over B because..." not "prefers A"

\textbf{Evidence-first and Less-is-more:}\\
- Extract patterns only with clear evidence (time/counts/quotes/details). Otherwise, skip.\\
- Do not hypothesize; do not fill gaps for completeness.\\
- If evidence for contrast is weak, state objective frequency only without conclusion.

\textbf{Output Standards:}\\
- English third-person, 300-500 words\\
- Structured paragraphs, focus on pattern recognition\\
- Each pattern includes: description + cases + uniqueness + frequency

\textbf{Input:}\\
- Historical weekly reports: \{previous\_weekly\_memories\}\\
- Current week: Year \{year\}, Week \{week\_number\} (\{week\_start\} to \{week\_end\})\\
- This week's daily reports: \{daily\_summaries\}

Generate weekly report directly:
\end{promptbox}

\subsubsection{L5 Profile Memory Consolidator}
\label{app:prompt_timem_consolidator_l5}
\noindent
\begin{promptbox}[L5 Profile Consolidator]
\promptbody
You are the agent's deep character profiler. Task: Build separate in-depth personalized archives for two participants, enabling the agent to clearly understand and distinguish their uniqueness.

\textbf{Core Objectives:}\\
Build deep personalized profiles, excavating each person's unique personality traits, behavioral patterns, and value orientations, enabling the agent to:\\
- Understand each person's uniqueness\\
- Distinguish differences between the two\\
- Predict likely behavioral choices\\
- Understand motivations behind decisions

\textbf{Analysis Positioning:}\\
L5 memory is the profiling layer, distilling individual core traits and uniqueness from long-term observations. Each person's archive should embody their irreplaceable personalized characteristics.

\textbf{Archive Structure (generate separately for each person):}\\

1. Basic Identity\\
Only list information explicitly mentioned in dialogue:\\
- Role positioning (gender identity / occupation / student, etc.)\\
- Life background (family status / geographic location, etc.)\\
- Main social contacts (who interacted with this month, frequency)

2. Key Events This Month\\
Select 3-5 events that best reveal this person's traits:\\
- Time + action + details + result\\
- Prioritize events involving choices, decisions, responses

3. Core Trait Analysis (focus, determines profile quality)

Identify 2-3 most unique traits for each person. Each trait must include:

[Trait Name] Use behavioral description, not adjectives

- Behavioral Evidence: At least 3 specific examples\\
\hspace*{1em}Time 1: What done + what said\\
\hspace*{1em}Time 2: What done + what said\\
\hspace*{1em}Time 3: What done + what said

- Uniqueness Explanation:\\
\hspace*{1em}Typical people's approach: [conventional choice]\\
\hspace*{1em}This person's approach: [unique choice]\\
\hspace*{1em}Resulting difference: [specific outcome]

- Supporting Quote: At least 1 direct quote

- Possible Underlying Motivation: Values or beliefs inferred from observations

4. Decision-Making Pattern\\
If decision scenarios exist, analyze:\\
- What prioritized in decisions\\
- Unique aspects of selection criteria\\
- Supporting case examples

5. Changes This Month\\
- Newly emerged behaviors or topics\\
- Shifts in attitude or frequency\\
- Specific description: From what to what

6. Comparative Distinction (Important!)\\
Clearly mark differences between the two:\\
- Characteristics both share\\
- Characteristics only this person has\\
- Characteristics contrasting with the other person

\textbf{Wording Standards:}\\
Use specific behavioral descriptions instead of abstract evaluations:\\
- "consults 2-3 people before each decision, considers for 3 days" not "cautious"\\
- "proactively shared experiences 4 times this month" not "open"\\
- "chose A because it helps X people" not "has a sense of mission"

\textbf{Key Requirements:}\\
- Each trait must have at least 3 examples\\
- Each trait must explain difference from typical people\\
- Two archives can clearly distinguish each other\\
- Support with data (counts, frequency, duration)

\textbf{Output Standards:}\\
- English third-person, 400-600 words per person\\
- Clear paragraphs, separate two archives with blank line\\
- Focus on trait analysis and difference identification

\textbf{Input:}\\
- Historical archives: \{previous\_monthly\_memories\}\\
- Current month: \{month\_name\} \{year\} (\{month\_start\} to \{month\_end\})\\
- This month's weekly reports: \{expert\_memories\}

Generate deep profiles separately for both participants:
\end{promptbox}

\subsubsection{Recall Gating Prompt for Simple Queries}
\label{app:prompt_timem_relevance_simple}
\noindent
\begin{promptbox}[Recall Gating (Simple Queries)]
\promptbody
Filter memories for simple fact query (Complexity 0).

Strategy: Aggressive filtering - Keep only direct answers\\
Target: 3-8 memories

\textbf{\#\# Filtering Rules}\\
1. KEEP if memory directly answers the question\\
2. KEEP if memory provides essential context (time/location of the fact)\\
3. EXCLUDE if related but does not contribute to answer\\
4. EXCLUDE if different topic entirely

\textbf{\#\# Instructions}\\
- Be strict: Only keep memories that help answer the specific question\\
- Remove noise: Exclude tangentially related memories\\
- Aim for 3-8 memories total

Question: \{question\}\\
Candidate memories (\{total\_count\} total):\\
\{numbered\_memories\}

Return IDs to keep (JSON format):\\
\{\{"relevant\_ids": [1, 2, 3, ...]\}\}
\end{promptbox}

\subsubsection{Recall Planner Prompt}
\noindent
\begin{promptbox}[Recall Planner]
\promptbody
You are a professional query intent analysis expert. Please select the most appropriate retrieval method based on the question type, and extract keywords.

\textbf{Critical Judgment Principles:}\\
- If the question requires understanding the user's preferences, habits, values, personality traits, or historical behavior patterns to answer correctly, classify as "Deep Retrieval" (2)\\
- If the question involves reasoning, prediction, evaluation, subjective judgment, or hypothetical scenarios, classify as "Deep Retrieval" (2)\\
- If the question requires integrating behaviors across multiple time points, multiple choices, or long-term trends to answer, classify as "Deep Retrieval" (2)\\
- Only single factual queries (who, when, where, what specific action) should be classified as "Simple Retrieval" (0)

\textbf{Retrieval Type Definitions:}

\textbf{0 - Simple Retrieval (Factual Questions):}\\
- Questions answerable by retrieving a single fact fragment\\
- Characteristics: Explicit time, location, person, event, or other objective fact queries\\
- Examples: "Where does X work?" "When did X go to location Y?" "Which meeting did X attend?"\\
- Key: Answer is an explicitly recorded fact, no reasoning or preference judgment needed

\textbf{1 - Hybrid Retrieval (Multi-Fact Integration Questions):}\\
- Questions requiring integration of multiple fact memories to answer\\
- Characteristics: Need to enumerate, summarize, or compare multiple facts, but no deep reasoning required\\
- Examples: "What activities did X participate in?" "What topics did X and Y discuss?" "Where has X been?"\\
- Key: Need to aggregate multiple facts, but still objective information integration

\textbf{2 - Deep Retrieval (Personalized Reasoning Questions):}\\
- Questions requiring reasoning based on user's deep personalized information (preferences, habits, values, personality) to answer\\
- Core Characteristics:\\
\hspace*{1em}* Need to understand user's stable preferences (what they like/dislike, values, interests)\\
\hspace*{1em}* Need to infer user's future behavior or likely choices ("Would like...?" "Suitable for...?" "Would choose...?")\\
\hspace*{1em}* Need to evaluate or judge user's personality traits, behavior patterns, cognitive style\\
\hspace*{1em}* Involves subjective judgment, evaluation, recommendation, prediction, hypothetical questions\\
\hspace*{1em}* Need to infer user's attitude or tendency based on historical behavior patterns\\
- Examples: "Would X enjoy a beach vacation?" "Is X an extroverted person?" "Might X be interested in programming?" "Does X prioritize career or family more?"\\
- Key: Answer requires synthesizing user's deep traits and preferences, not directly recorded facts

\textbf{Judgment Process:}\\
1. First identify: Does the question require user's preferences/habits/personality/values? If yes $\rightarrow$ Deep Retrieval (2)\\
2. Second identify: Does the question require reasoning/prediction/evaluation/subjective judgment? If yes $\rightarrow$ Deep Retrieval (2)\\
3. Third identify: Does the question require summarizing multiple fact fragments? If yes $\rightarrow$ Hybrid Retrieval (1)\\
4. Finally: If only single explicit fact needed $\rightarrow$ Simple Retrieval (0)

Keyword extraction requirements:\\
1. Extract 1-3 most important keywords from the question\\
2. Exclude common stopwords (such as: the, a, in, is, have, and, or, with, etc.)\\
3. STRICTLY FORBIDDEN: Never include any personal names, usernames, or names\\
4. FOCUS ONLY ON: Action words, object names, location types, concept words, adjectives and other non-name key concepts

Question: \{question\}

Please carefully analyze the essential needs of the question and output in the following JSON format:\\
\{\textbackslash n\\
\hspace*{1em}"complexity": 0/1/2,\textbackslash n\\
\hspace*{1em}"keywords": ["keyword1", "keyword2", "keyword3"]\textbackslash n\\
\}
\end{promptbox}